%% file: root.tex
\documentclass[letterpaper, 10 pt, conference]{ieeeconf}  
\input{imports}

\IEEEoverridecommandlockouts                              

\title{\LARGE \bf
Detecting and Mitigating System-Level Anomalies of Vision-Based Controllers}

\author{Aryaman Gupta$^{*1}$, Kaustav Chakraborty$^{*2}$, Somil Bansal$^{2}$
\thanks{*The first two authors contributed equally to this paper.}
\thanks{$^{1}$Author is with the ECE Department at the Indian Institute of Technology (BHU), Varanasi. {\tt\footnotesize aryaman.gupta.ece20@itbhu.ac.in.}}
\thanks{$^{2}$Authors are with the ECE Department at the University of Southern California. {\tt\footnotesize \{kaustavc, somilban\}@usc.edu.}
This research is supported in part by the DARPA Assured Neuro Symbolic Learning and Reasoning (ANSR) program and by the NSF CAREER program (2240163). Aryaman was supported by the IUSSTF-Viterbi Summer Research Program.}
}

\begin{document}

\maketitle
\thispagestyle{empty}
\pagestyle{empty}

\begin{abstract}
\input{abstract}
\end{abstract}

\section{Introduction}
\label{sec:intro}
\input{intro_new} 

\section{Problem Setup}
\label{sec:problem}
\input{problem}
\section{Background: Hamilton-Jacobi Reachability}
\label{sec:background}
\input{background}

\section{Learning and Mitigating System Failures using Reachability Analysis }
\label{sec:approach}
\input{approach}

\section{Case Study: Autonomous Aircraft Taxiing}
\label{sec:cases}
\input{taxinet}


\section{Discussion and Future Work}
\label{sec:conclusion}
\input{conclusion}




\bibliographystyle{IEEEtran}
\bibliography{./references}

\end{document}

%% file: imports.tex
\usepackage[Symbolsmallscale]{upgreek}
\usepackage{xargs}
\usepackage{float}
\usepackage{wrapfig}
\usepackage[normalem]{ulem}
\usepackage{amsmath,amsfonts}
\usepackage{algorithm}
\usepackage{array}
\usepackage{textcomp}
\usepackage{stfloats}
\usepackage{url}
\usepackage{verbatim}
\usepackage{graphicx}
\usepackage{cite}
\usepackage{xcolor}

\usepackage{graphicx}
\usepackage{caption}
\usepackage{subcaption}
\usepackage{svg}
\usepackage{algorithm, setspace}
\usepackage[noend]{algpseudocode}
\usepackage[%
    colorlinks=true,
    pdfborder={0 0 0},
    linkcolor=red
]{hyperref}

\algnewcommand\algorithmicforeach{\textbf{for each}}
\algdef{S}[FOR]{ForEach}[1]{\algorithmicforeach\ #1\ \algorithmicdo}

\usepackage{graphics} 
\usepackage{epsfig} 
\usepackage{mathptmx} 
\usepackage{times} 
\usepackage{amsmath} 
\usepackage{amssymb}  
\usepackage{textcomp, gensymb}
\usepackage{url}
\usepackage{graphicx}
\usepackage{color}
\usepackage{xcolor}
\usepackage{soul}
\usepackage{mathtools}
\usepackage{algorithm, setspace}
\usepackage[noend]{algpseudocode}
\usepackage{ccicons}
\usepackage{comment}
\usepackage{bbm}
\usepackage{multirow}

\newcommand{\state}{\mathbf{x}}
\newcommand{\dyn}{f}

\newcommand{\ctrl}{u}

\newcommand{\obs}{\mathcal{O}}

\newcommand{\policy}{\uppi}
\newcommand{\sensor}{S}
\newcommand{\image}{I}

\newcommand{\traj}[2]{\zeta^{#1}_{#2}}

\newcommand{\brt}{\mathcal{V}}
\newcommand{\ibrt}{\mathcal{I}_{unsafe}}
\newcommand{\inprod}[2]{{\left\langle #1, #2 \right\rangle}}

\newcommand{\classy}{\sigma}


%% file: abstract.tex
Autonomous systems, such as self-driving cars and drones, have made significant strides in recent years by leveraging visual inputs and machine learning for decision-making and control. Despite their impressive performance, these vision-based controllers can make erroneous predictions when faced with novel or out-of-distribution inputs. Such errors can cascade to catastrophic system failures and compromise system safety. In this work, we introduce a run-time anomaly monitor to detect and mitigate such closed-loop, system-level failures. Specifically, we leverage a reachability-based framework to stress-test the vision-based controller offline and mine its system-level failures. This data is then used to train a classifier that is leveraged online to flag inputs that might cause system breakdowns.
The anomaly detector highlights issues that transcend individual modules and pertain to the safety of the overall system. We also design a fallback controller that robustly handles these detected anomalies to preserve system safety. 
We validate the proposed approach on an autonomous aircraft taxiing system that uses a vision-based controller for taxiing. 
Our results show the efficacy of the proposed approach in identifying and handling system-level anomalies, outperforming methods such as prediction error-based detection and ensembling, thereby enhancing the overall safety and robustness of autonomous systems.\\
Website: {\tt\footnotesize phoenixrider12.github.io/FailureMitigation}

%% file: intro_new.tex
With the advances in deep learning and computer vision, modern autonomous and robotic systems have reached a level of competence that, in some instances, exceeds human capabilities \cite{kaufmann2023champion}. 
Nevertheless, given the vast array of scenarios these vision-driven controllers might face in the real world, we can never entirely rule out the occurrence of uncommon corner cases and failure scenarios.
Thus, even as we aspire for our robots to adapt to new conditions, there is a growing need for runtime anomaly detection systems that can provide early alerts when a system encounters anomalies, helping to counteract and mitigate potential rare failures \cite{rahman2021run}.

Anomaly detection (AD) methods for learning components typically fall under two categories: distributional shift methods and functional uncertainty methods \cite{sinha2022system}.
The former aims to detect and mitigate distribution shifts between training and test times.
These methods include approaches that artificially inject noise in the ground truth data (e.g., images) to force a distribution shift and create anomalous data \cite{sun2021complementing}. 
Another line of research in this direction aims to develop learning algorithms that are distributionally robust, optimizing the worst-case performance within a pre-specified envelope of distributional shifts to guarantee out-of-distribution performance \cite{Ben-TalHertogEtAl2013, DuchiNamkoong2021} or performing domain randomization during training \cite{tobin2017domain}.
Since detecting distribution shifts can be challenging in general, especially when the test distribution is unknown \textit{a priori}, the functional uncertainty methods instead detect the inputs that are either dissimilar to the training data \cite{SalehiMirzaeiEtAl2021,RuffKauffmanEtAl2021, YangZhouEtAl2021, narayanan2018learning, ji2022proactive} or lead to erroneous or low-confidence predictions \cite{AbdarPourpanahEtAl2021, LakshminarayananPritzelEtAll2017, GalZoubin2016, zhang2014predicting, SharmaAzizanEtAl2021, AminiSchwartingEtAl2020}.
The above methods predominately detect
anomalous inputs at the component level; however, such
component level monitoring (e.g., detecting image
classification errors) can be insufficient to prevent
system-level faults.
Seemingly minor errors in the individual modules can cascade into catastrophic effects at a system level. 
Hence, a system-level view of such problems is often encouraged and is the main motivation behind this work.

\begin{figure}[t]
\centering
\includegraphics[width=\columnwidth]{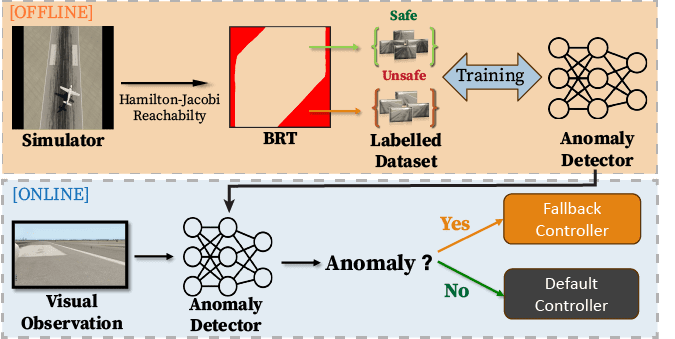}
\caption{\small{Detecting and mitigating system-level anomalies of vision based controller. \textit{\textbf{Offline} (top row):} We compute the Backward Reachable Tube (BRT) of the vision-based system using Hamilton-Jacobi Reachability analysis \cite{chakraborty2023discovering}. This allows us to create a labeled dataset of safe and unsafe images without any manual intervention. We train the anomaly detector using this dataset. \textit{\textbf{Online} (bottom row):} We run the system on previously unseen environments where the trained anomaly detector triggers a fallback controller or the default controller depending on whether the observed image is anomalous.}}
\vspace{-2.25em}
\label{fig:flowdiagram}
\end{figure}

In this work, we present an approach to detect and mitigate such system-level anomalies for autonomous systems that leverage learning-driven vision-based controllers for decision making (Fig. \ref{fig:flowdiagram}).
Our work builds upon the study conducted in \cite{chakraborty2023discovering} that leverages a visual simulator within a Hamilton-Jacobi Reachability framework to expose the closed-loop failures of the system under the vision-based controller.
However, the proposed framework requires privileged information about the test environment and is computationally not suitable for online applications.
Our key idea is to utilize this framework \textit{offline} to stress-test and automatically mine the system-level failures of the closed-loop system across a variety of environment conditions.
Ultimately, this process provides us with a diverse set of input image sequences which when seen by the vision-based controller leads to an overall system failure.
These images are then used to train a simple classifier that can serve as an anomaly detector during runtime.
We demonstrate that the resultant AD implicitly leverages the revealed failure modes to classify whether a previously unseen input image has the potential to trigger a system failure, without requiring any domain-specific heuristics.
Next, we employ the AD to trigger a simple fallback controller online that can trade the system's performance to assure system safety whenever the anomaly detector determines the system is at risk of entering the failure zone. 
We illustrate the joint properties of our anomaly detector and fallback controller on an autonomous aircraft taxiing case study, leveraging a vision-based controller. 
We compare our method against commonly used component-level AD techniques, such as predicting ensemble-based uncertainty, and prediction error-based anomaly detectors to highlight the key advantages of the proposed system-level anomaly detector.  

%% file: problem.tex
%
Consider a robot with dynamics, $\dot{\state} = \dyn(\state, \ctrl)$
%
%
where, state $\state \in \mathbb{R}^n$ and control $\ctrl \in \mathbb{U}$ (a compact set). The robot possesses a sensor $\sensor$, that allows it to perceive visual inputs from its surroundings at any given state $\state$.  We can express the mapping between the state to the input as $I = \sensor(\state)$. $\image$ could be an RGB image, a pointcloud, etc. 
Additionally, we have a vision-based controller, $\policy$, that maps $\image$ to the control $\ctrl$, and defined as, $\ctrl \coloneqq \policy(\image)$.
Note that $\policy$ can be an end-to-end policy, or it can consist of several sub-modules, some of which may not be data-driven.

Let $\traj{\policy}{\state}(\tau)$ be the robot's state achieved at time $\tau$ when it starts from state $\state$ at time $t=0$, and follows the policy $\policy$ over $[0,\tau]$. Finally, let $\obs$ denote a set of undesirable states (or failure states) for the system. 
As an example, $\obs$ could represent obstacles for a ground robot.
Thus, an initial state $\state$ is considered unsafe for the system if $\exists s\in [0,\tau],\,  \traj{\policy}{\state}(s) \in \obs$. 
The set of input images the system sees, starting from such unsafe states, are thus considered \textit{anomalous for the closed-loop system} (as they eventually steer the system to $\obs$) and denoted as $\ibrt$.

In this work, we are primarily interested in obtaining a mapping, $\classy$, that provides a binary decision of whether a given input $\image$ can possibly lead to the failure of the system: 
%
\vspace{-0.4em}
\begin{equation}
    \classy: \image \rightarrow \ \{0,1\}
    \label{eqn:classifier_statement}
    \vspace{-0.4em}
\end{equation}
where $1$ means that $\image$ is anomalous, and $0$ means it is not.
Hence, an ideal $\classy$ should output $1$ whenever $\image \in \ibrt$ and $0$ otherwise.
In addition, we seek to find a mitigation system that can preserve system safety even if the robot encounters an anomalous input in $\ibrt$.

Generating $\classy$ requires addressing a few challenges: (a) $\ibrt$ is typically hard to obtain as the test environment is unknown \textit{a priori}; (b) even in known environments, obtaining high-dimensional inputs, e.g., RGB images, that lead to system-level failures is a challenging problem; finally, (c) for vision-based controllers, $\ibrt$ varies with changes in the robot's surroundings, e.g., $\ibrt$ for an indoor navigation robot might change as we move from one room to another. 
Thus, formulation of a general AD pipeline that can recognize features of anomalous input images without prior knowledge of the new surroundings is non-trivial.

\begin{figure}[h]
\centering
\includegraphics[width=\columnwidth]{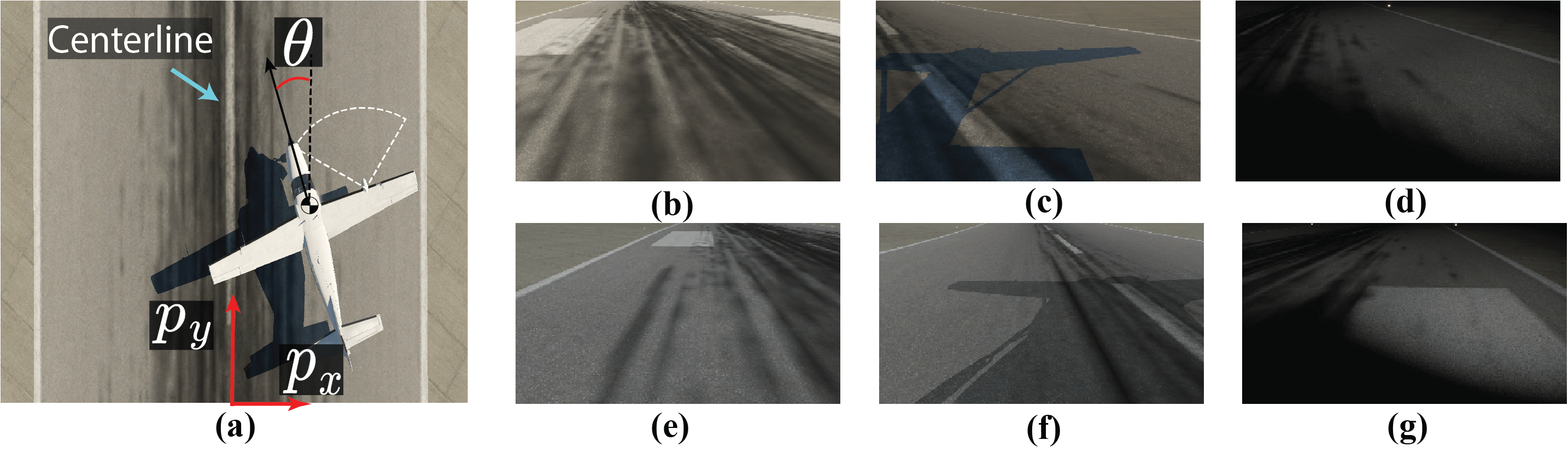}
\caption{\small{\textbf{(a)} $p_{x}$, $p_{y}$, $\theta$ denote the state of the aircraft; dashed-white lines show FoV of the camera. Runway simulation images with clear sky at \textbf{(b)} 9AM, \textbf{(c)} 5PM, and \textbf{(d)} 9PM, and overcast clouds at \textbf{(e)} 9AM, \textbf{(f)} 5PM, and \textbf{(g)} 9PM taken by camera mounted on the aircraft for the KMWH runway showing the variations in lighting conditions and shadows (\textbf{(c),(f)}) for changes in the environment.}}
\label{fig:taxinet}
\end{figure}

\vspace{0.5em}
\noindent \textbf{\textit{Running example (TaxiNet).}} We introduce the autonomous aircraft taxiing problem \cite{katz2021verification} as a running example to illustrate the key aspects of our framework.
Here, the robot is a Cessna 208B Grand Caravan aircraft modeled as a three-dimensional non-linear system with dynamics:
%
\vspace{-0.8em}
\begin{equation}
\label{eqn:dyn_taxinet}
     \dot p_x=v\, sin(\theta)\quad \dot p_y=v\, cos(\theta)\quad  \dot \theta = u
     \vspace{-0.6em}
\end{equation}
where $p_x$ is the crosstrack error (CTE), $p_y$ is the downtrack position (DTP) and $\theta$ is the heading error (HE) of the aircraft in degrees from the centreline (Fig. \ref{fig:taxinet}(a) shows how these quantities are measured). $v$ is the linear velocity of the aircraft kept constant at 5 m/s, and the control $u$ is the angular velocity. 
\vspace{-1em}
\vspace{-0.75em}

The goal of the aircraft is to follow the centreline as closely as possible using the images obtained through a camera mounted on its right wing. 
For this purpose, the aircraft uses a Convolutional Neural Network (CNN), which returns the estimated  CTE and HE, $(\hat p_x, \hat \theta)$. A proportional controller (P-Controller) then takes these predicted tracking errors to return the control input as follows:
\vspace{-0.6em}
\begin{equation}
    \label{eqn:pctrl}
    \ctrl \coloneqq tan(-0.74\hat p_x -0.44\hat \theta)
    \vspace{-0.6em}
\end{equation}
Hence, the policy $\policy$ is a composition of the CNN and the P-Controller.
Intuitively, the P-controller is designed to steer the aircraft towards the centreline based on the state estimate provided by the CNN \cite{katz2021verification}.  We are interested in analyzing this \textit{given} controller without modifying its underlying design.

The image observations are obtained using the X-Plane flight simulator that can render the RGB image, $I$, from a virtual camera ($\sensor$) mounted on the right wing of the aircraft at any state and a given time of day (see Fig. \ref{fig:taxinet}(b)-(g) for representative images under different simulation conditions). 

We define the unsafe states for the aircraft as $\mathcal{O} = \{\state: |p_x| \geq B\}$, where $B$ is the runway width. Thus, $\obs$ corresponds to aircraft leaving the runway.
%
%
Our goal is to find the mapping, $\classy$, that is able to classify whether an input image $\image$ is likely to eventually drive the system off the runway.

%% file: background.tex
In this work, we will use Hamilton-Jacobi (HJ) Reachability analysis to obtain a dataset of system-level anomalies offline.
We now provide a brief overview of HJ reachability and refer the readers to \cite{bansal2017hamilton} for more details.

In reachability analysis, we focus on calculating the \textit{Backward Reachable Tube (BRT)} of the system. 
The BRT refers to the collection of initial states from which an agent following policy $\policy(\state)$, can reach the target set $\obs$ within the time interval $[t, T]$:
%
\vspace{-0.6em}
\begin{equation}
\vspace{-0.6em}
\mathcal{V} \coloneqq \{\state: \exists \tau \in [t, T], \zeta^{\policy}_{\state}(\tau) \in \mathcal{O} \}
\label{eqn:brt}
\end{equation}

HJ reachability analysis allows us to compute the BRT for general nonlinear systems.
To compute the BRT, the target set is first represented as a sub-zero level set of a function $l(\state)$, denoted as $\obs = \{\state: l(\state) \leq 0\}$ \cite{mitchell2005time,mitchell2002level}. 
The function $l(\state)$ typically represents the signed distance from a state to the target set $\obs$. With this formulation, the BRT computation can be reframed as an optimal control problem that involves finding a value function defined as:
\vspace{-0.6em}
\begin{equation}
V(\state,t) = \min_{\tau \in [t, T]} l(\zeta^{\policy}_{\state}(\tau))
\vspace{-0.6em}
\label{eqn:vfn}
\end{equation}

This value function (defined in \eqref{eqn:vfn}), can be iteratively computed using dynamic programming principles leading to a partial differential equation known as the Hamilton-Jacobi-Bellman Variational Inequality (HJB-VI) \cite{bansal2017hamilton}:
\vspace{-0.5em}
\begin{equation}
\begin{aligned}
\min\{D_tV(\state,t)+H(\state,t)&,l(\state) - V(\state,t)\}=0 \\
\text{with }V(\state,T) = l(\state)
\label{eqn:hjivi}
\vspace{-1em}
\end{aligned}
\end{equation}
In this equation, $D_t$ and $\nabla$ represent the time and spatial gradients of the value function, respectively. The Hamiltonian, denoted as $H \coloneqq \inprod{\nabla V(\state,t)}{f(\state, \policy(\state)}$, embeds the system dynamics in the HJI-VI.
Consequently, the BRT corresponds to the subzero level set of the value function:
\vspace{-0.5em}
\begin{equation}
\mathcal{V} = \{\state: V(\state,t) \leq 0\}
\vspace{-0.5em}
\end{equation}
In the next section, we describe how we can use $\mathcal{V}$ for obtaining a system-level anomaly detector.

%

%% file: approach.tex
%
%
%
%
In this work, we use a learned classifier model as our anomaly detector. 
Our key idea is to compute the BRT of the system offline under a vision-based controller for a diverse set of environments. 
The BRT can then be used to label the training data for our classifier.
We now describe this process in detail over three steps.
%
%
%
%
\subsection{Automatic labels for system-level anomalous inputs}
To compute a dataset of anomalous inputs, we first compute the set of all starting states that lead the closed-loop system to $\obs$ under the vision-based controller.
In other words, we compute the BRT of the system under the vision-based controller.
Once the BRT is obtained, the images corresponding to the states inside the BRT can be used as examples of system-level anomalous inputs, i.e.,  
\vspace{-0.5em}
\begin{equation}
\image \in \ibrt \Leftrightarrow \state \in \brt, \text{ where},\, \image = \sensor(\state)
\label{eqn:failure_brt}
\vspace{-0.5em}
\end{equation}
%

However, the BRT computation in \eqref{eqn:hjivi} typically requires an analytical model of $\pi(x)$, which is not possible in our case, since an analytical model of vision sensor $S$ is generally not available. 
To overcome this challenge, we will follow the approach used in \cite{chakraborty2023discovering}, wherein the BRT is computed using image samples collected from a photorealistic simulator.
These image samples are then processed through the vision-based controller to obtain samples of $u$, which are subsequently used to approximate the BRT.
Thus, access to photorealistic simulators provides us with an opportunity to inexpensively sample from high-dimensional state spaces, which could otherwise be a challenging and tedious process. 
Moreover, since the image observed at a particular state depends on the environment conditions (e.g., lighting conditions, weather, etc.), we compute BRTs for a diverse set of environment conditions to capture a wider set of system anomalies in our training dataset, allowing for a better generalization during runtime.

Once all the BRTs are computed, we can sample states randomly in the state space, render the corresponding images, and automatically label any sampled image as safe or anomalous depending on whether the state is outside or inside the BRT.
Note that by the construction of the training dataset, our method targets image inputs that lead to system-level failures, without requiring any manual labeling.

\subsection{Learning an anomaly detector}
%
Equipped with a training dataset of system-level anomalies, we train a binary classifier to predict the label for a novel input image.
Specifically, we use a deep neural network (DNN)-based classifier, given their remarkable success with image classification tasks.
The DNN takes as input an image and returns the softmax scores of the input being an anomaly. 
During training, we assume access to the system's BRT and hence the true labels of an image being anomalous.
However, during testing, we do not possess the BRT information. Our model is seen to identify anomalies during testing without needing the BRT by learning the underlying features of anomalous inputs. We evaluate the performance and highlight some interesting cases of the detected anomalies by the trained classifier in Sec. \ref{sec:cases}.

\subsection{Fallback controller}
Equipped with an anomaly detector, we propose a safety-preserving controller pipeline for the system. 
Specifically, when the system detects a possible anomalous input, we switch the system's vision-based controller $\policy$ with a fallback controller $\policy^*$. 
For an input $\image$, the overall control input for the system is given as:
%
\vspace{-0.5em}
\begin{equation}
\vspace{-0.25em}
    \ctrl_{\text{filtered}}= 
\begin{cases}
    \policy^*(\image)
    ,& \text{if } \classy(\image) = 1\\
    \policy(\image),              & \text{otherwise,}
\end{cases}
\end{equation}
where $\classy$ is the learned anomaly detector.
The condition $\classy(\image) = 1$ is triggered if $\image$ is an anomaly, and hence the system switches to $\policy^*$. 

The choice of $\policy^*$ is primarily aimed at maintaining system safety and may vary from system to system. For example, $\policy^*$ might be an alternative controller that is safer but computationally intensive (e.g., a full-stack SLAM for navigation) or trade-off performance for safety (e.g., coming to a complete stop), or logistically expensive (e.g., relying on a human operator). Designing a good $\policy^*$ for a system is itself an interesting research question that we defer to future work.

%% file: taxinet.tex
%
\noindent \textbf{\textit{Generation of the Anomaly Dataset.}}
To generate data for training the AD, we compute BRT of the aircraft under the TaxiNet controller across 
two different times of the day (9 AM and 9 PM), two cloud conditions (clear and overcast), and three different runways (codenamed: KMWH, KATL, PAEI), leading to 12 different BRTs.
Note that each of the above conditions might lead to a different visual input $I$ at the same state $\state$ (see Fig. \ref{fig:taxinet} (b)-(e) for samples images), and hence lead to a different control input being applied to the aircraft, and subsequently, different BRTs. 

To compute the BRTs, we use the Level Set Toolbox (LST)\cite{mitchell2007toolbox} that solves the HJB-VI in \eqref{eqn:hjivi} numerically over a state-space grid for a time horizon of 8s. 
Following \cite{chakraborty2023discovering}, we use a uniform $101\times101\times101$ grid over  $p_x \in [-X,X]m$, $p_y \in [100,250]m$ and $\theta \in [-28\degree, 28\degree]$ (the value of $X$ depends on the runway, e.g., for KMWH, X = 11).
The LST requires the control input $u$ at each of the grid points to compute the BRT, which is obtained by rendering the image using X-Plane simulator at that state and querying the controller in \eqref{eqn:pctrl}.
This consists of passing the rendered image through the TaxiNet CNN, followed by a P-controller.
We refer the readers to \cite{chakraborty2023discovering} for more details on the BRT computation.

\begin{figure}[ht]
\centering
\includegraphics[width=\columnwidth]{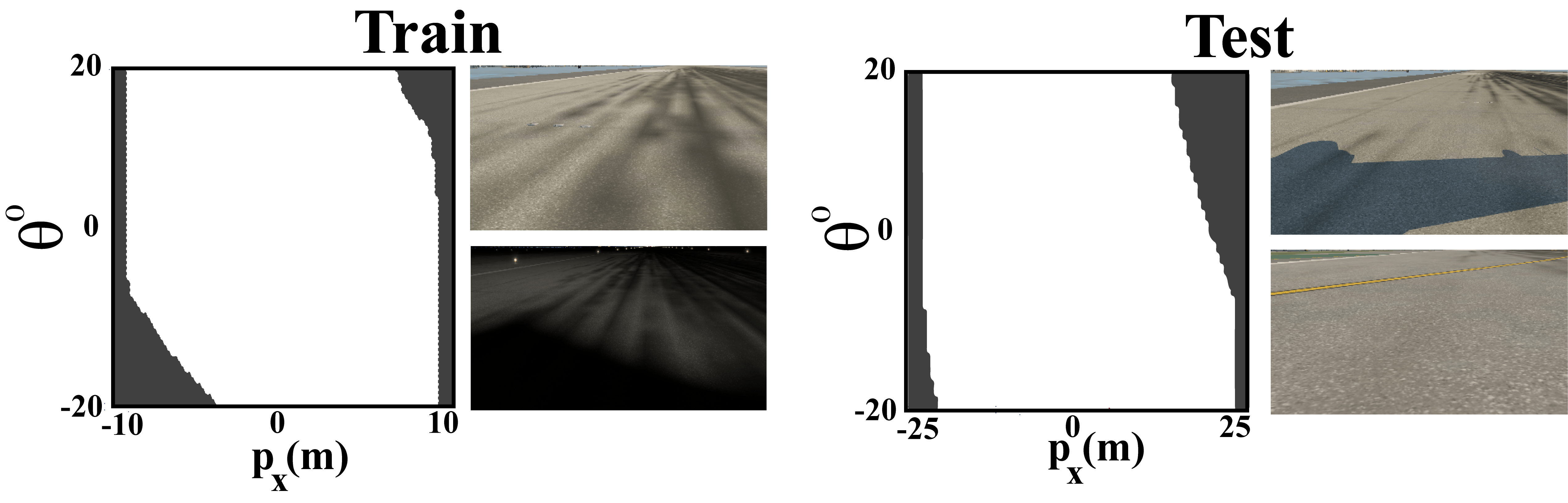}
\caption{\small{BRT of 9AM KMWH runway (part of train set) shown on the left along with a few training images and BRT of 5PM KSFO runway (part of test set) shown on the right, along with a few testing images, showing diversity in our training and testing scenarios.}}
\vspace{-1.25em}
\label{fig:brt_samples}
\end{figure}

%
Fig. \ref{fig:brt_samples} (left) shows a slice of the computed BRT for the KMWH runway at 9am in clear conditions for $p_y = 100m $.  
The gray area represents the set of starting states of the aircraft from which it will eventually leave the runway under TaxiNet, whereas the white represents the safe area.
We next randomly sample 20K states for each of the 12 conditions and render the images for these states, leading to an overall training dataset of 240K images.
If an image is generated from a state present in the BRT, we label it as an anomaly. Otherwise, we label it as safe.
%
\vspace{-0.75em}

\noindent \textbf{\textit{Anomaly Detector.}} 
We next train a binary classifier on the collected dataset that takes a 224x224x3 input image and returns the probability of the image being anomalous. 
We use a pre-trained EfficientNet-B0 Model \cite{tan2019efficientnet} as our backbone, replacing the last layer with a fully connected layer that feeds into a binary softmax output layer.
We trained the classifier using cross-entropy loss and Adam optimizer for 20 epochs with a fixed learning rate of $3e^{-4}$, which took around 1 hour on a Nvidia RTX 3090 GPU.

\begin{figure*}[t]
\centering
\includegraphics[width=\textwidth]{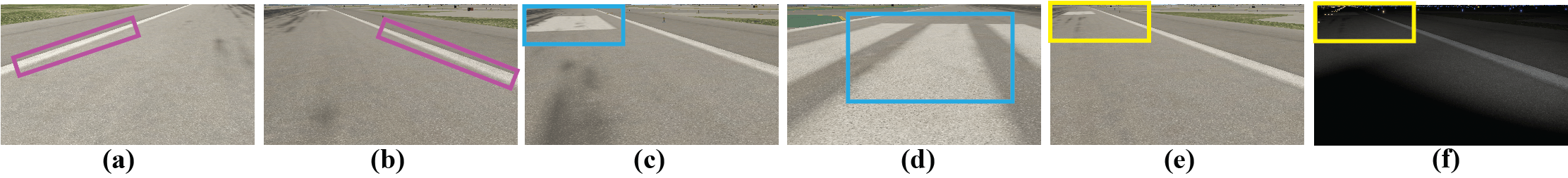}
\vspace{-0.75em}
\caption{\small{Failures detected by AD. \textbf{(a, b)} Images correspond to the aircraft being close to the runway boundaries (highlighted with the magenta bounding boxes).\textbf{(c, d)} The visual controller confuses the runway markings  (highlighted with the cyan bounding boxes) with the centerline and ultimately leads to a system failure. \textbf{(e, f)} Image (f) is (accurately) not classified as an anomaly during the night time (the same image is classified as anomaly during the day, shown in (e)), as the runway lights (highlighted with the yellow bounding boxes) help the visual controller to predict its position accurately and thereby avoid failure.}}
\vspace{-1.75em}
\label{fig:all_failures}
\end{figure*}

\vspace{0.5em}
\noindent \textbf{\textit{Evaluation.}} 
To test the generalization capabilities of the learned AD, we test it at an unseen time of the day (5 PM) for the three airports present in the training set, as well as on two unseen KSFO and KEWR airports, across both cloud conditions (clear and overcast). 
Specifically, we measure the recall and accuracy of the learned AD. 
A higher recall value tells us that our AD can reliably detect true positives, i.e., the AD returns a conservative prediction if it is unsure of an input. 
Such a behavior err on caution to produce a safety-first detector. 
To compute these metrics, we also obtain the BRT of the system for the test environments; however, note that during testing, our network has access to only the sampled image \textit{without} any privileged information regarding the system, surroundings, or the BRTs. 
\input{table2}
We summarize the performance of the anomaly detector in Table \ref{table:table2} (Here (C) = cloudy, (O) = overcast).
As evident from the results, the learned AD is consistently able to detect system-level anomalies in new environments. 
We illustrate some of the representative images that were classified as anomalous in Fig. \ref{fig:all_failures}. 
The AD is able to learn the safe limits of the runway and that if the system is close to the runway boundary. TaxiNet may not be able to correctly estimate aircraft's state and it might fail under the vision-based controller.
This may not be particularly surprising as most states near the runway boundary (white lines in Fig. \ref{fig:all_failures} (a-b)) are unsafe in the training dataset (the region near the runway boundary is contained in the BRT in Fig. \ref{fig:brt_samples}), potentially allowing the AD to learn that such images often lead to a failure.
However, what's interesting is that the AD learns that a similar image \textit{will not} cause a system failure during night time because of the runway lights, which, when are lit, help the TaxiNet to estimate the aircraft position more accurately, avoiding a failure.  
Finally, we noticed that several images classified as anomalies contained the aircraft runway markings (Fig. \ref{fig:all_failures} (d-f)). Such an emergent understanding of semantic failure mode is quite impressive since in the prior work \cite{chakraborty2023discovering}, the authors had to manually analyze the failures to recognize these patterns as anomalies. 
Our AD justifies the use of a learned classifier by practically automating the detection process without any inclusion of heuristics or manual intervention.

\vspace{0.5em}
\noindent {\textbf{\textit{Baselines.}}}
We next compare our method to a few commonly used techniques for anomaly detection.
\subsubsection{Prediction Error-based Labels}
\begin{figure}[H]
\centering
\includegraphics[width=0.8\columnwidth]{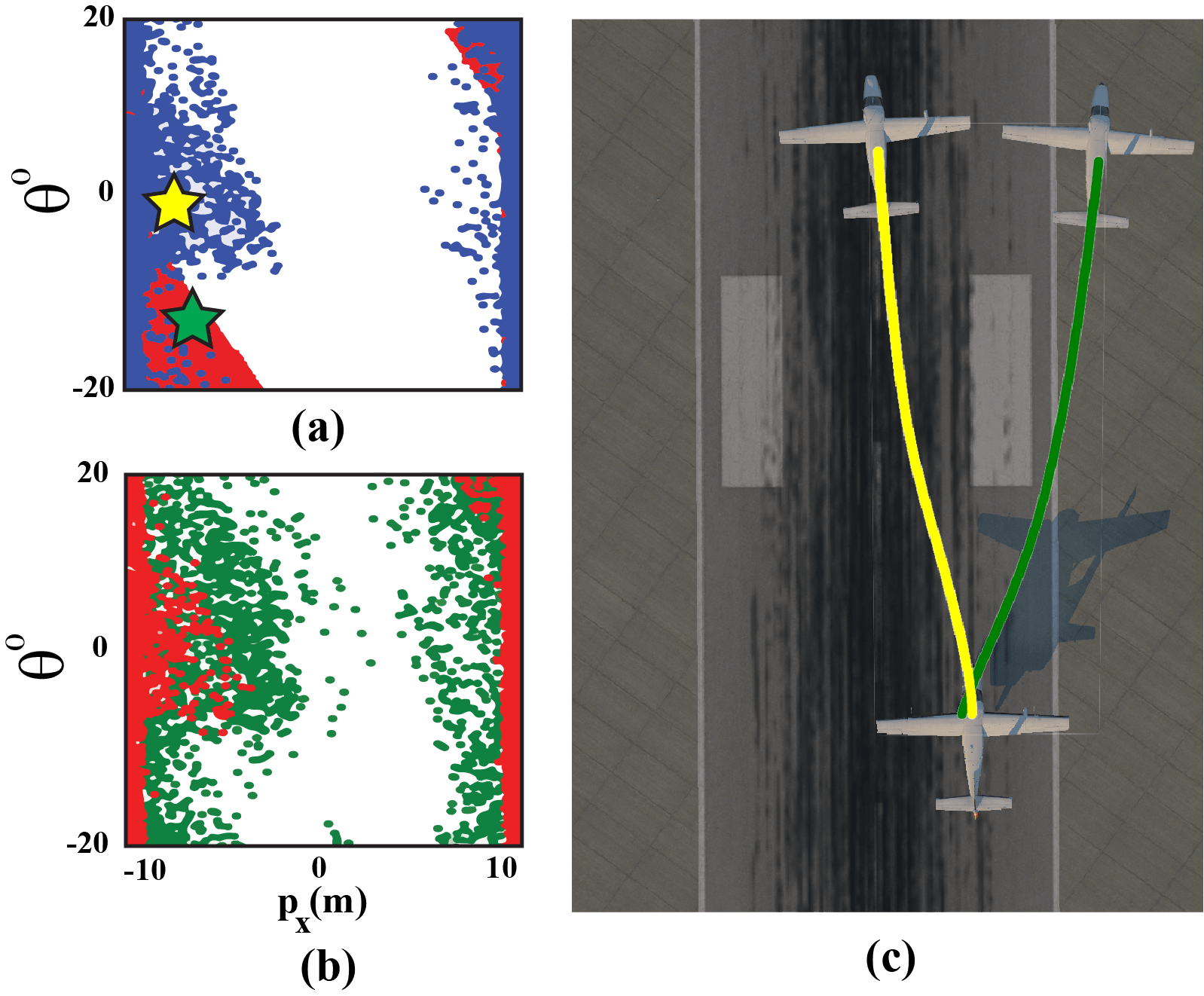}
\captionof{figure}{\small{\textbf{(a)} Comparison between prediction error (blue) and BRT-based (red) labels. \textbf{(b)} Prediction error-based labels for $threshold=0.3$ (green) and $threshold=0.6$ (red). \textbf{(c)} Yellow and Green lines show trajectories starting from the yellow and green stars, respectively.}}
\vspace{-1.25em}
\label{fig:error_baseline}
\end{figure}

We use a prediction error-based labeling scheme instead of the proposed HJ Reachability-based scheme to collect the training dataset for our AD. 
Specifically, if the TaxiNet prediction error is above a certain threshold for a particular image, then we label this image as an anomaly.  
We overlay the prediction error-based anomaly labels (blue) on top of the BRT-based labels (red) for one of the environments in Fig. \ref{fig:error_baseline}(a).
\vspace{-0.75em}
It is evident that the prediction error-based labels may not be a good representative of the system-level failures. 
For example, the states near the Yellow star are anomalous as per prediction error but do not actually cause the system failure (the yellow trajectory in Fig. \ref{fig:error_baseline}(c)), resulting in a pessimistic AD and hampering the system performance. 
On the other hand, certain states and images may have a small error from the TaxiNet module perspective (states near the Green star) and are not classified as anomalies; yet, they cascade to a system failure (the green trajectory in Fig. \ref{fig:error_baseline}(c)).
This results in an overly optimistic nature of the prediction of error-based AD in certain regions.
This ``unpredictable'' nature of the prediction error-based AD persists with the change in the error threshold used for labeling (Fig. \ref{fig:error_baseline}(b)).
Unsurprisingly, we observe the same phenomenon in the AD trained on these prediction-error-based labels and, thus, omit those results for brevity.
This leads us to conclude that component-level, prediction error-based labels may not be a good representative for determining system-level failures. 

\subsubsection{Ensembled Predictions}
Another popular mechanism to detect anomalies is based on the predictive uncertainty of an ensemble of neural networks.
To design an ensemble, we train 5 different versions of the TaxiNet with different weight initializations. 
\begin{wrapfigure}{r}{0.35\columnwidth}
          \centering
          \vspace{-1.5em}
         \includegraphics[width=0.35\columnwidth]{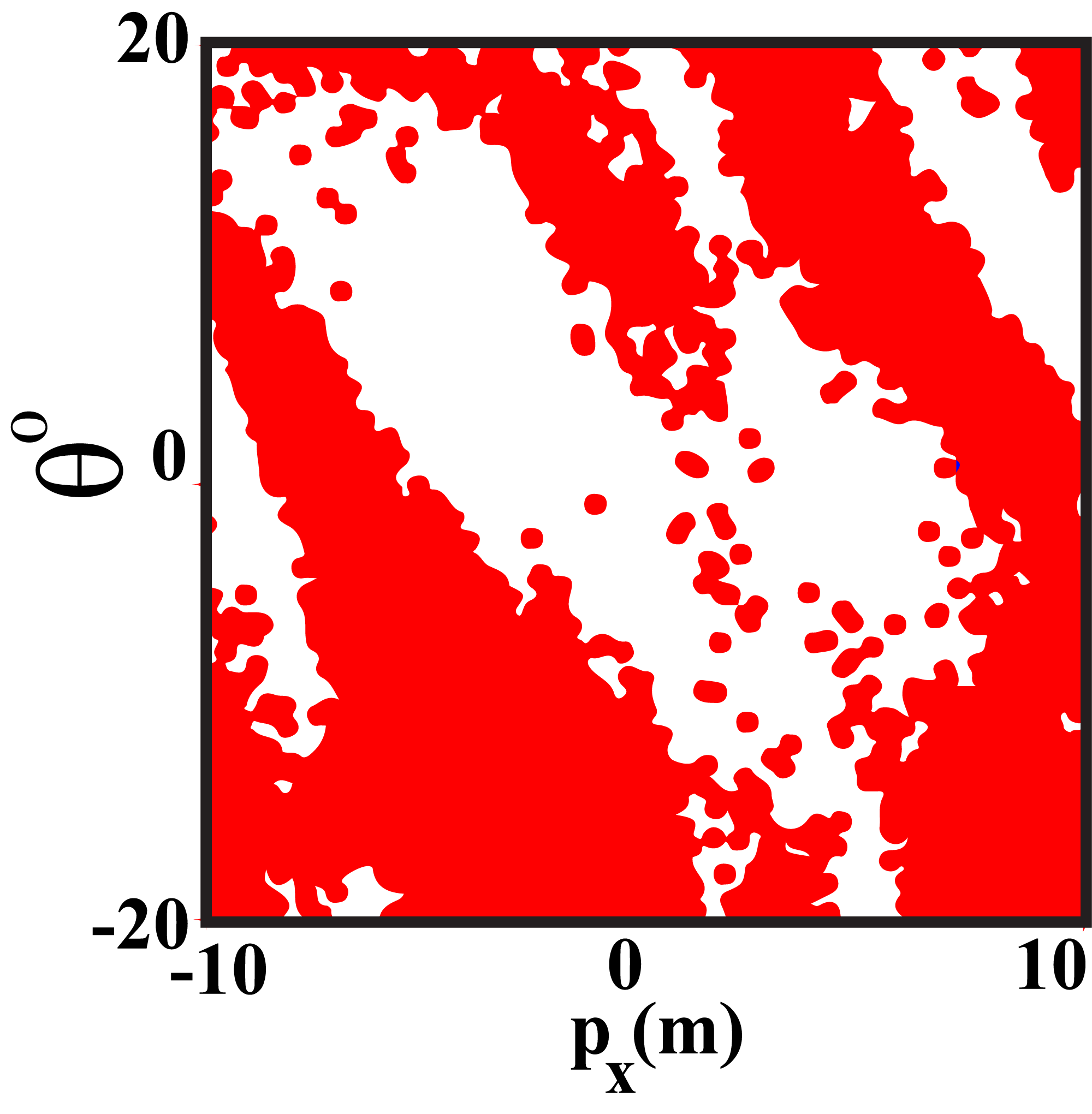}
         \vspace{-1.5em}
         \caption{\small{Labels generated using ensembling denoting failures (red) and success (white).}}
         \vspace{-1.5em}
         \label{fig:ensemble}
\end{wrapfigure}
If the variance between the predictions exceeds a threshold for any input image, we assign such an image as an anomaly. 
The corresponding anomalies over the statespace are shown in Fig. \ref{fig:ensemble}. 
We observe that this method does not perform well because the ensemble confidently makes incorrect predictions for some states (around the top right of the statespace), leading to faulty labels in those states. 
On the other hand, for some states (near the central region of the statespace), the ensemble disagreed on the predictions, leading to states being incorrectly marked as an anomaly. 
Hence, an ensembling-based approach also fails to accurately predict system-level anomalies.

\begin{figure}[h!]
     \centering         \includegraphics[width=0.8\columnwidth]{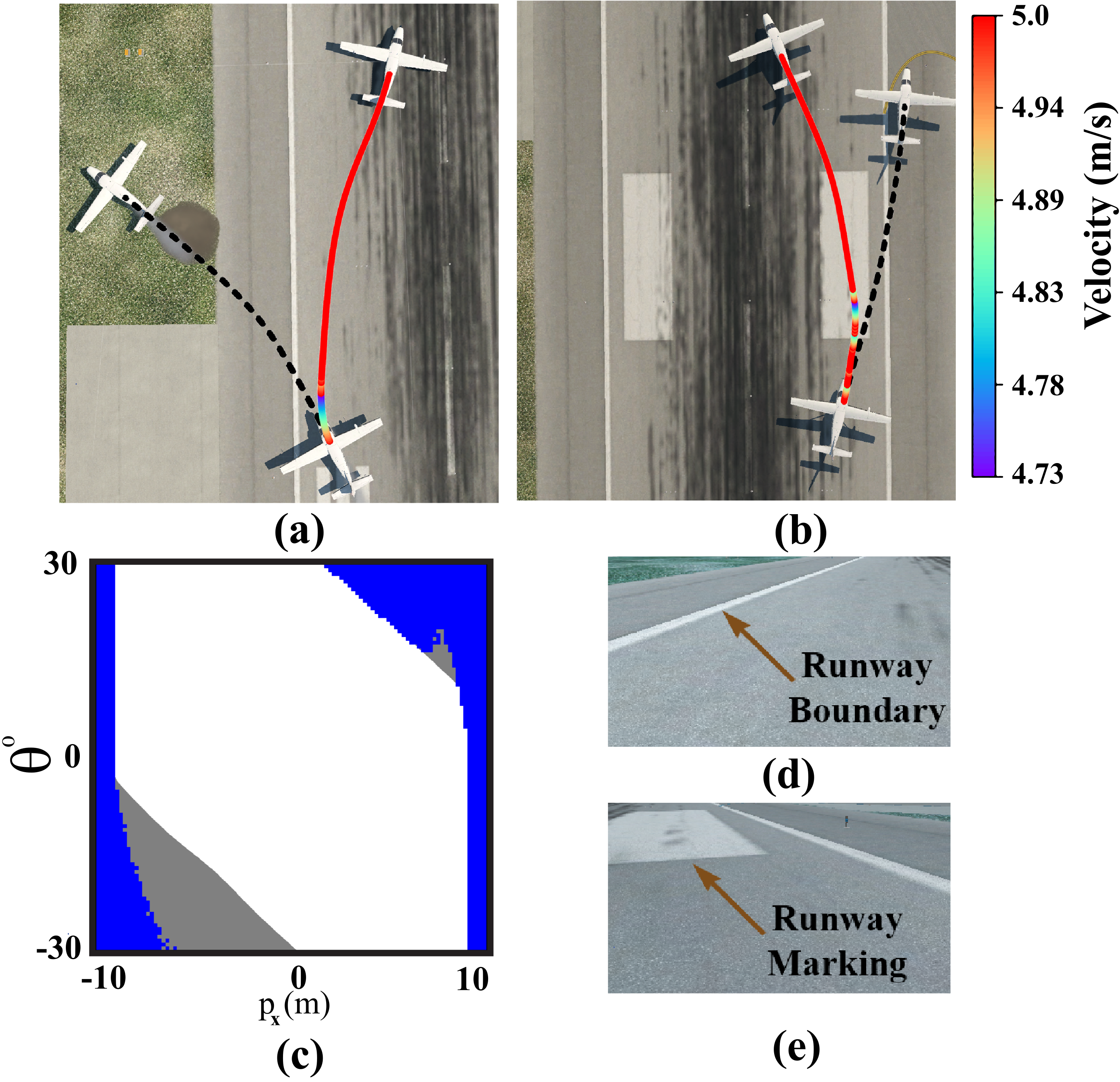}
     \vspace{-0.75em}
\caption{\small{\textbf{(a, b)} Trajectory followed by the aircraft under the TaxiNet controller (dashed black line) and the safety pipeline (red line). The color shift in the red curve shows velocity variation due to the fallback controller. \textbf{(c)} The grey region represents the system BRT under the TaxiNet controller, and the blue region represents the BRT under the safety pipeline. The BRT obtained using the AD and the fallback controller is appreciably smaller than the one obtained using vanilla TaxiNet. \textbf{(d)} Input image at the start state in (a), causing system failure due to runway boundaries. \textbf{(e)} Input image at the start state in (b), causing system failure due to runway markings.}}
\vspace{-1.0em}
\label{fig:good_results}
\end{figure}

\vspace{0.5em}
\noindent {\textbf{\textit{Fallback Mechanism.}}}
Equipped with a capable AD, we designed a simple fallback mechanism to ensure our aircraft's safety under anomalous inputs. If at any point in its trajectory, the aircraft observes an image $\image$ that is classified as an anomaly (i.e., $\classy(I) = 1$), the linear velocity of the plane ($v$ in Eqn. \eqref{eqn:dyn_taxinet}) is reduced by $0.01 m/s$. Intuitively, this results in slowing the aircraft every time it encounters an anomaly, ultimately coming to a complete stop if it continues to encounter anomalies. As soon as the aircraft detects an image that is non-anomalous, the default TaxiNet controller takes over. 
Fig. \ref{fig:good_results} shows trajectories of the aircraft under the resultant controller (red trajectories). 
The nominal trajectories without the fallback controller are shown in dashed black.
In the first case (Fig. \ref{fig:good_results} (a)), the anomaly is triggered due to the aircraft approaching too close to the runway boundary, while in the second case (Fig. \ref{fig:good_results} (b)) the anomaly is triggered due to semantic failure of the runway markings.
In both cases, the TaxiNet controller leads the system off the runway.
On the other hand, the fallback controller decreases the aircraft velocity whenever an anomaly is triggered to ensure system safety. 
This can also be seen from the velocity variation along the red trajectories. 
To further illustrate the advantage of the fallback controller, we compute the system BRT under the Taxinet controller and the fallback controller pipeline for one of the test cases on KMWH runway (Fig.\ref{fig:good_results}(c)).
The BRT (i.e., the set of failure states) is much smaller under the safety pipeline (the BRT volume decreases from 25.75\% to 18.24\%), showing that the proposed mechanism significantly reduces the number of closed-loop system failures.
\vspace{-0.75em}
Finally, with the inclusion of the AD and the fallback controller along with the TaxiNet CNN, the system was able to run at 25Hz as compared to 33Hz when the TaxiNet governs the system as the sole controller. This allows us to run the system online without any severe latency issues. 

\vspace{-0.75em}
\begin{figure}[h]
     \centering     \includegraphics[width=0.8\columnwidth]{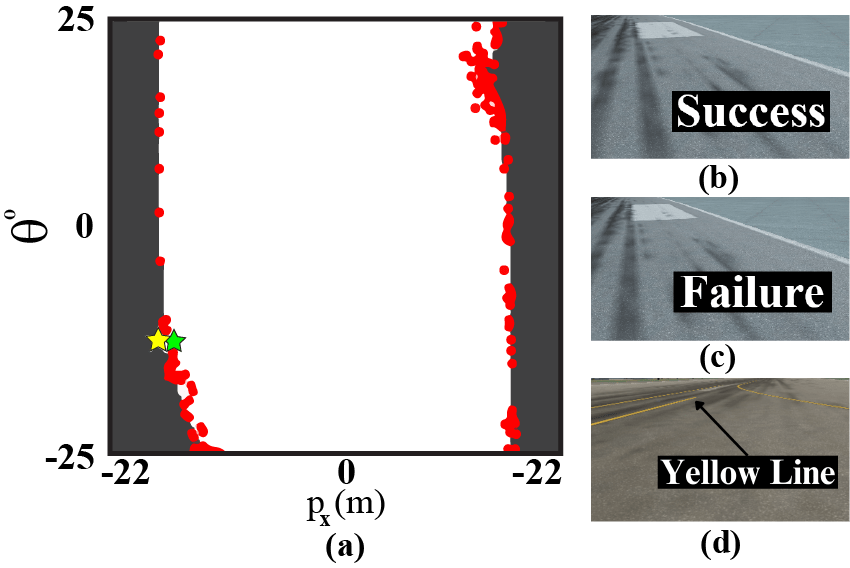}
     \vspace{-0.5em}
\caption{\small{\textbf{(a)} Red dots represent incorrect predictions of the AD. The incorrect predictions are concentrated around the BRT boundary. 
\textbf{(b)} Input image corresponding to the green star in (a), incorrectly classified as failure by AD.  
\textbf{(c)} Input image corresponding to the yellow star in (a), incorrectly classified as success by AD. 
\textbf{(d)} Test runway contains yellow-colored lines that confuse the TaxiNet, but such anomalies are not predicted well, as no runway with yellow lines is present in the training dataset.
\vspace{-1em}
}}
\label{fig:bad_results}
\end{figure}
\noindent \textbf{\textit{Failure Modes of AD.}} 
Even with all these exciting discoveries, the AD fails to detect some anomalies. 
We find that this happens particularly for the images corresponding to the states near the BRT boundary. 
For example, in Fig. \ref{fig:bad_results}(b) and \ref{fig:bad_results}(c), we show two visually similar images in our test dataset (corresponding to the green and yellow stars in Fig. \ref{fig:bad_results}(a)).
Even though the two images are visually similar, one image is inside the BRT and leads to the system failure (hence an anomaly), while the other does not. 
Such similar images with minor differences are hard to detect for our AD. Finally, we noticed that some of the semantics of the test environment that cause anomalies are not present in the training dataset and are not predicted well by the proposed AD (Fig. \ref{fig:bad_results}(d)). Such issues highlight the need for a continual update of anomaly detectors as more data about the system anomalies is obtained. 

\begin{figure}[H]
\centering
\vspace{-0.75em}
\includegraphics[width=0.75\columnwidth]{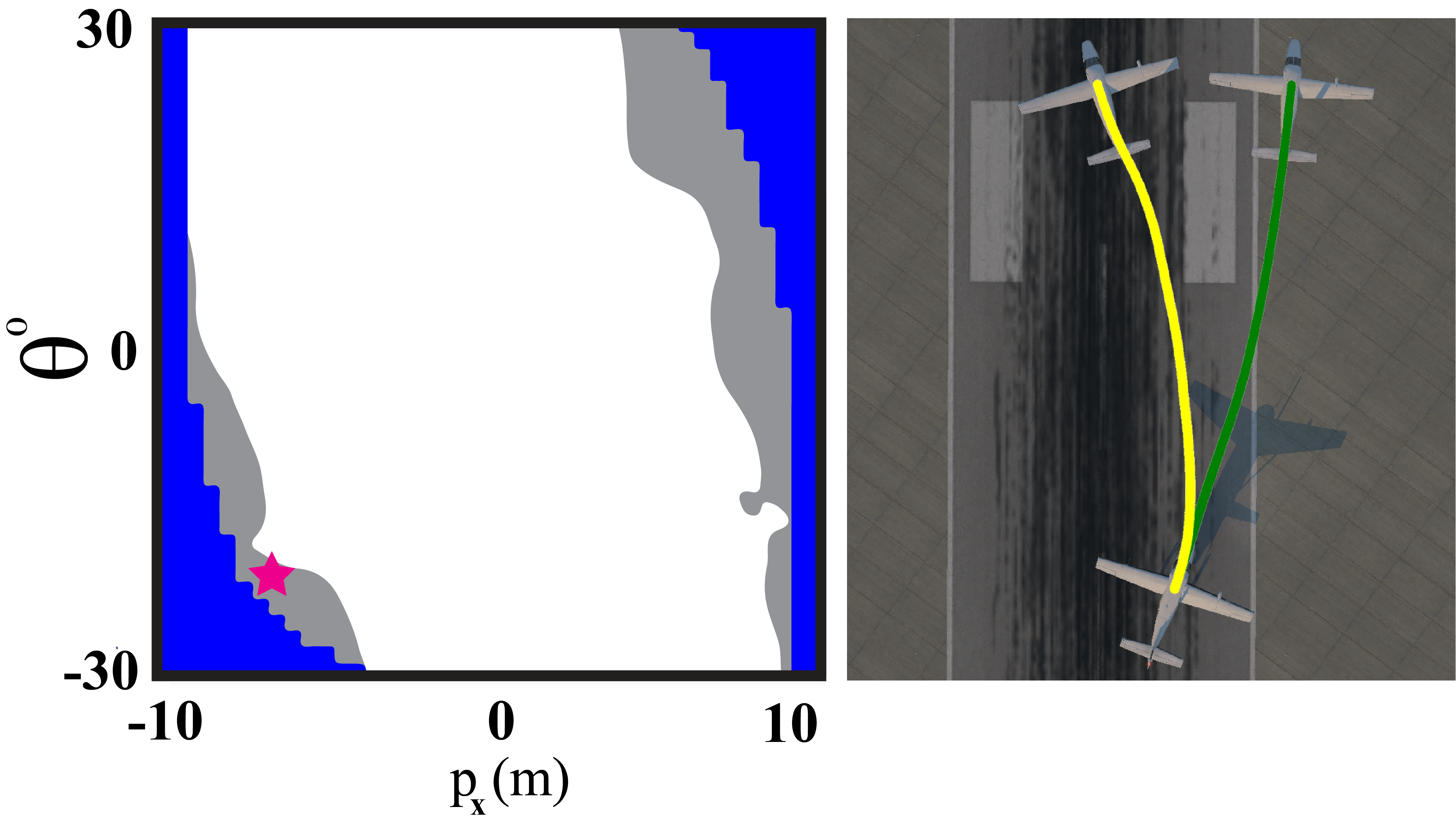}
\vspace{-0.75em}
\caption{\small{\textbf{(Left)} BRT slices for a DTP of 100m at 5 PM 
 on KMWH runway showing the improvements due to incremental training (blue) over the vanilla TaxiNet (grey). The state depicted by the magenta star is only included in the grey region. \textbf{(Right)} Trajectory simulated from this state is seen to succeed under the incrementally trained TaxiNet (yellow) but fails under the vanilla version (green).}}
 \vspace{-1.25em}
\label{fig:incremental}
\end{figure}

\vspace{0.5em}
\noindent \textbf{\textit{Incremental Training.}} A complementary approach to employing an AD can be to perform targeted incremental training of the TaxiNet model on the collected anomalous data. 
Essentially, these labeled anomalies represent scenarios where TaxiNet fails to perform optimally. Therefore, by conducting specialized training on these failure instances, we aim to fortify TaxiNet's robustness in handling such cases.
We present some preliminary results of incremental training of TaxiNet in Fig. \ref{fig:incremental}. We plan on exploring this method further in future work.
%
%

%% file: table2.tex
\begin{table}[htbp]
\centering
\caption{Performance of the learned anomaly detector.}
\begin{tabular}{ccccc}
\hline
\rule{0pt}{1\normalbaselineskip}
\multirow{2}{*}{\textbf{Airport ID}} & \multicolumn{2}{c}{\textbf{Recall (\%)}} & \multicolumn{2}{c}{\textbf{Accuracy (\%)}} \\
                                     & (C)                 & (O)                & (C)                  & (O)                 \\[1mm]
\hline
\rule{0pt}{1\normalbaselineskip}
KMWH                                 & 91.50               & 92.22              & 95.89                & 97.02               \\
KATL                                 & 97.44               & 95.89              & 98.19                & 97.97               \\
PAEI                                 & 93.24               & 93.71              & 97.71                & 97.99               \\
KEWR                                 & 96.31               & 94.20              & 96.29                & 96.72               \\
KSFO                                 & 90.34               & 90.56              & 89.72                & 90.62           \\[1mm]
\hline   
\label{table:table2}
\end{tabular}%
\vspace{-1em}
\end{table}

%% file: conclusion.tex
In this work, we present an approach aimed at identifying and mitigating system-level anomalies of autonomous systems that rely on vision-based controllers for decision-making. By leveraging insights from reachability analysis, our approach learns an anomaly detector that effectively tackles concerns related to system-level safety during runtime. The learned anomaly detector is combined with a fallback controller to reduce the potential for catastrophic system failures significantly. In the future, we plan to test our approach over a wide range of real-world scenarios and focus on more intricate fallback mechanisms that can efficiently work for different system-level failures. In addition, to further enhance the performance of the AD, one can take advantage of temporal information, such as the visual history, instead of only the current image. Finally, instead of relying on a classifier, other approaches could explore unsupervised methods like clustering via support vector machines or k-means applied to our annotated dataset. 

We introduce a framework for automatically discovering the closed-loop failures of vision-based controllers. Our work combines simulation-based approaches with HJ reachability analysis to systematically and tractably find these failures. We demonstrate the efficacy of our approach on two distinct applications -- a 5-dimensional wheeled robot and a 3-dimensional aircraft using RGB-image-based neural network controllers. Our work suggests a number of interesting future directions. First, the grid-based approach to computing reachable sets suffers from the curse of dimensionality. In the future, it would be interesting to consider alternative sampling-based reachability methods, such as DeepReach \cite{bansal2021deepreach}, that are shown to scale well to high-dimensional systems. We would also like to explore using the obtained failures in improving the performance of the vision-based controller, e.g., through incremental training on the obtained scenarios. Finally, even though the failure discovery is automatic, the analysis of the failures is primarily performed manually in the current work. Automating the failure analysis, e.g., through generative and clustering methods, will be a promising future direction.